\documentclass[11pt]{article}
\PassOptionsToPackage{sort}{natbib}
\usepackage[preprint]{acl}
\usepackage{booktabs}
\usepackage{multirow}
\usepackage{enumitem}
\usepackage[utf8]{inputenc}

\definecolor{authority}{HTML}{6A4C93}
\definecolor{care}{HTML}{00B4A6}
\definecolor{fairness}{HTML}{3498DB}
\definecolor{loyalty}{HTML}{E74C3C}
\definecolor{sanctity}{HTML}{F39C12}

\definecolor{clrPhase}{HTML}{F0F0F0}

\usepackage{mdframed}
\usepackage{xcolor}
\usepackage{listings}

\newmdenv[
  backgroundcolor=clrPhase,
  linecolor=black!15,
  linewidth=0.4pt,
  roundcorner=4pt,
  innerleftmargin=8pt,
  innerrightmargin=8pt,
  innertopmargin=6pt,
  innerbottommargin=6pt
]{annotatorbox}

\lstdefinestyle{pytorch}{
    language=Python,
    basicstyle=\ttfamily\tiny,
    keywordstyle=\color{violet}\bfseries,
    stringstyle=\color{teal},
    commentstyle=\color{gray}\itshape,
    numberstyle=\tiny\color{gray},
    numbers=none,
    numbersep=6pt,
    frame=single,
    rulecolor=\color{gray!40},
    breaklines=true,
    showstringspaces=false,
    tabsize=4,
    xleftmargin=1em,
    framexleftmargin=0.8em,
    backgroundcolor=\color{gray!8},
    morekeywords={self,None,True,False}
}
\usepackage{times}
\usepackage{latexsym}
\usepackage{algorithm}
\usepackage{algpseudocode}
\usepackage{listings}
\usepackage{tikz}
\usetikzlibrary{shapes,arrows,arrows.meta,fit,positioning,shadows}
\usetikzlibrary{bayesnet}
\usetikzlibrary{external}
\usepackage{adjustbox}
\usepackage{svg}
\usepackage{fontawesome5}
\usepackage{tcolorbox}
\usepackage{orcidlink}
\usepackage[T1]{fontenc}
\usepackage[utf8]{inputenc}
\usepackage{microtype}
\usepackage{caption}
\usepackage{subcaption}
\usepackage{booktabs}
\usepackage{multirow}
\usepackage{inconsolata}
\usepackage{amsfonts}
\usepackage{amssymb}
\usepackage{amsmath}
\usepackage[nameinlink]{cleveref}

\usepackage[draft]{fixme}

\newcommand{\chip}[3]{%
  \tikz[baseline=-0.5ex]{\node[fill=#1!12, draw=#1!55, line width=0.4pt,
    rounded corners=2pt, inner sep=1.8pt, font=\tiny\bfseries, text=#1!45!black] {#2\ #3};}%
}
\newcommand{\ebar}[2]{%
  \tikz[baseline=-0.15ex]{
    \draw[rounded corners=1pt, fill=black!7, draw=none] (0,0) rectangle (0.9,0.1);
    \draw[rounded corners=1pt, fill=#1!65, draw=none] (0,0) rectangle ({#2*0.9},0.1);
  }%
}
\newcommand{\careicon}{\faHandsHelping}
\newcommand{\fairicon}{\faBalanceScale}
\newcommand{\loyicon}{\faShield*}

\title{Moral Entropy: Auditing Bias and Uncertainty in Moral Judgment}

\author{Maciej Skorski\ \orcidlink{0000-0003-2997-7539} \\
  University of Luxembourg \\
  \texttt{maciej.skorski@gmail.com} \\}

\begin{document}
\maketitle

\begin{abstract}
Most work in computational ethics treats annotator disagreement on moral content as noise to be voted away, collapsed into majority vote or the more permissive \emph{any-annotator} rule the moment a single annotator flags an item. We argue this uncertainty should instead be modeled and learned from. We introduce \textbf{Moral Entropy}, a Bayesian framework that keeps a full posterior over the true label and decomposes its entropy into \emph{aleatoric} uncertainty (irreducible disagreement about the moral content) and \emph{epistemic} uncertainty (from insufficient or noisy annotation) --- and lets any heuristic consensus rule be audited against a calibrated ground truth via entropy methods such as cross-entropy/KL, Brier score, and expected calibration error. Across three corpora and fifteen discourse domains, auditing the standard aggregation rules against this posterior reveals bias that no current pipeline reports: the any-annotator rule disagrees with the calibrated posterior on roughly $30\%$ of items --- pooled, almost entirely false positives, though the errors invert at the foundation level ($19.9\%/38.9\%$ mean $\mathrm{FPR}/\mathrm{FNR}$ on MFTC) --- while the stricter majority and two-vote rules miss $63$--$83\%$ of true positives.

\textbf{Keywords:} Computational Ethics, Moral Foundations Theory, Bayesian modeling, calibration, proper scoring rules, annotator disagreement
\end{abstract}

\section{Introduction}

\subsection{Motivation}

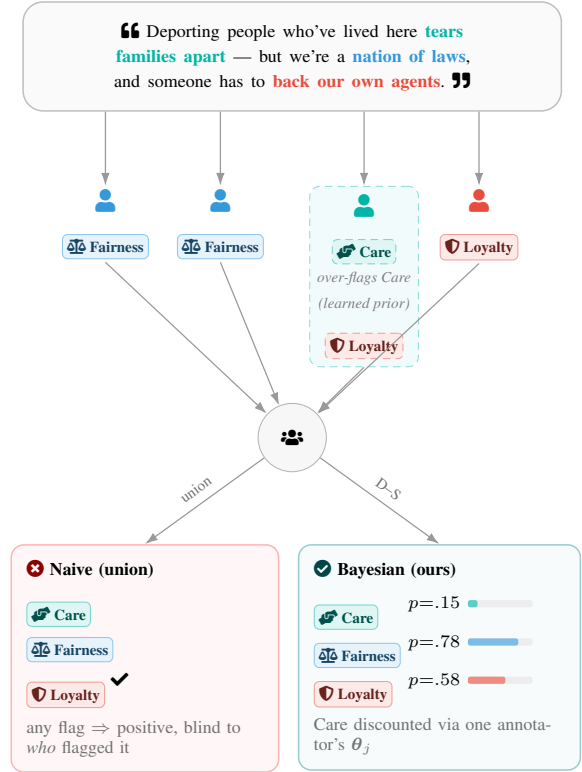
\begin{figure}[t!]
\centering
\begin{adjustbox}{max width=0.98\columnwidth}
\begin{tikzpicture}[
  every node/.style={font=\small},
  annot/.style={align=center, inner sep=1pt, anchor=north},
  outcome/.style={draw=black!20, line width=0.6pt, rounded corners=5pt, align=left,
                   inner sep=6pt, text width=3.5cm, font=\scriptsize},
  bubble/.style={draw=black!25, line width=0.6pt, rounded corners=8pt, fill=black!2,
                 align=center, inner sep=8pt, text width=0.9\columnwidth},
  hub/.style={draw=black!30, circle, fill=black!3, minimum size=0.95cm, align=center, font=\scriptsize},
  arr/.style={-{Latex[length=1.8mm,width=1.2mm]}, line width=0.5pt, black!40},
  lbl/.style={font=\tiny, align=center, text=black!55}
]
 
\node[bubble] (tweet) at (0,0)
  {\scriptsize\faQuoteLeft\ Deporting people who've lived here
   \textcolor{care}{\textbf{tears families apart}} --- but we're a
   \textcolor{fairness}{\textbf{nation of laws}}, and someone has to
   \textcolor{loyalty}{\textbf{back our own agents}}.\ \faQuoteRight};
 
\node[annot] (c1) at (-2.6,-1.8)  {\textcolor{fairness}{\small\faUser}\\[2pt] \chip{fairness}{\fairicon}{Fairness}};
\node[annot] (c2) at (-1.0,-1.8)  {\textcolor{fairness}{\small\faUser}\\[2pt] \chip{fairness}{\fairicon}{Fairness}};
\node[annot, draw=care!45, densely dashed, rounded corners=4pt,
      inner sep=3pt, fill=care!5] (c3) at (1.0,-1.8)
  {\textcolor{care}{\small\faUser}\\[2pt] \chip{care}{\careicon}{Care}\\[2pt]
   {\tiny\itshape\color{black!50} over-flags Care}\\
   {\tiny\itshape\color{black!50} (learned prior)}\\[2pt] \chip{loyalty}{\loyicon}{Loyalty}};
\node[annot] (c4) at (2.6,-1.8)   {\textcolor{loyalty}{\small\faUser}\\[2pt] \chip{loyalty}{\loyicon}{Loyalty}};
 
\foreach \i in {1,...,4}{
  \draw[arr] (tweet.south -| c\i.north) -- (c\i.north);
}
 
\node[hub] (hub) at (0,-5.3) {\faUsers};
\foreach \i in {1,...,4}{
  \draw[arr] (c\i.south) -- (hub);
}
 
\node[outcome, fill=red!3, draw=red!30, below=1.0cm of hub.south, xshift=-2.05cm, text width=3.3cm] (naive)
  {{\color{red!55!black}\faTimesCircle}\ \textbf{Naive (union)}\\[4pt]
   \chip{care}{\careicon}{Care}\\[3pt]
   \chip{fairness}{\fairicon}{Fairness}\\[3pt]
   \chip{loyalty}{\loyicon}{Loyalty}\ \faCheck\\[4pt]
   {\color{black!55} any flag $\Rightarrow$ positive, blind to \emph{who} flagged it}};
 
\node[outcome, fill=teal!3, draw=teal!40, below=1.0cm of hub.south, xshift=2.05cm, text width=3.5cm] (bayes)
  {{\color{teal!55!black}\faCheckCircle}\ \textbf{Bayesian (ours)}\\[4pt]
   \renewcommand{\arraystretch}{1.2}%
   \begin{tabular}{@{}l@{\hspace{3pt}}r@{\hspace{3pt}}l@{}}
   \chip{care}{\careicon}{Care} & $p{=}.15$ & \ebar{care}{0.15} \\
   \chip{fairness}{\fairicon}{Fairness} & $p{=}.78$ & \ebar{fairness}{0.78} \\
   \chip{loyalty}{\loyicon}{Loyalty} & $p{=}.58$ & \ebar{loyalty}{0.58} \\
   \end{tabular}\\[3pt]
   {\color{black!55} Care discounted via one annotator's $\boldsymbol{\theta}_j$}};
 
\draw[arr] (hub) -- node[lbl, pos=0.5, sloped, above] {union} (naive.north);
\draw[arr] (hub) -- node[lbl, pos=0.5, sloped, above] {D--S} (bayes.north);
 
\end{tikzpicture}
\end{adjustbox}
\caption{One sentence, three foundations. Fairness and Loyalty are each flagged independently by two annotators, but Care by only one --- an annotator (c3) with a documented history of over-flagging Care, learned from their confusion matrix $\boldsymbol{\theta}_j$ across many other items. The naive union rule cannot tell the difference and flags all three alike. Our Bayesian consensus discounts Care in proportion to that annotator's known unreliability \emph{for that foundation specifically} ($p{=}.15$); it also mildly discounts Loyalty ($p{=}.58$) since one of its two corroborators is the same annotator, whose $\boldsymbol{\theta}_j$ is estimated per foundation and happens to be trustworthy for Loyalty; Fairness, corroborated by two otherwise-clean annotators, is discounted least ($p{=}.78$).}
\label{fig:graphical-abstract}
\end{figure}

Annotators disagree on moral judgments, and often for good reason: Moral Foundations Theory itself predicts systematic disagreement across ideological and cultural lines \citep{graham2013moral,haidtWhenMoralityOpposes2007}. Almost every MFT-annotated corpus destroys this signal before anyone downstream can use it. The standard way to turn raw annotator votes into a single training label is a heuristic consensus rule: majority vote, or the more permissive \emph{any-annotator} rule (positive, if flagged by any annotator) \citep{hooverMoralFoundationsTwitter2020,tragerMoralFoundationsReddit2022,hoppExtendedMoralFoundations2021,preniqiMoralBERTFineTunedLanguage2024,zangari-etal-2025-me2,bullaLargeLanguageModels2025}. Both are attractive because they are cheap and assumption-free. Both also discard exactly the information a moral-psychology researcher should care about most: how much annotators agreed, and why.

This matters for two distinct reasons. First, a dataset with high inter-annotator disagreement may be capturing a real moral phenomenon rather than noisy labeling --- MFT does not predict convergence. Second, aggregation rules like any-annotator are statistically biased in a way no one reports: because the rule is a logical OR over $J$ noisy binary votes, its false-positive rate grows mechanically with $J$ even when every individual annotator is well calibrated. A downstream classifier trained on such labels inherits this bias with no way to detect it, since the heuristic consensus reports no uncertainty by construction.
 
We argue both problems are solvable by a Bayesian consensus model, that yields a full posterior over the true label rather than a single bit. The entropy of that posterior is a natural, principled uncertainty measure --- what we call \textbf{Moral Entropy} --- and it decomposes cleanly into the two uncertainty types above. Once this posterior exists, any heuristic consensus rule can be audited against it directly, with standard tools from statistical decision theory (such as proper scoring rules).

\subsection{Contribution}

\textbf{Auditing consensus rules.} We introduce \textbf{Moral Entropy} --- the entropy of the Bayesian posterior over a moral-foundation label. Cross-entropy, Brier score, and calibration error are all Bregman divergences of an entropy function --- one family, many possible diagnostics; we instantiate the 0/1-loss case in the main text and the remainder in the supplement. Across three corpora ($106{,}627$ texts carrying over $250{,}000$ annotations), heuristic consensus rules disagree with the Bayesian posterior on a large, corpus-dependent share of items: substantial bias, whose direction depends on the rule.

\textbf{Entropy-based uncertainty analysis.} We decompose Moral Entropy into aleatoric (irreducible disagreement) and epistemic (model) uncertainty, revealing three qualitatively distinct regimes across corpora.

\textbf{Soft-label fine-tuning.} Training classifiers on the calibrated posterior instead of collapsed hard labels yields consistent 2--3\% accuracy gains, driven by the greater consistency of soft targets.

\subsection{Related Work}

\begin{table}[t]
\centering
\tiny
\begin{tabular}{@{}l p{0.76\columnwidth}@{}}
\toprule
\multirow{2}{*}{\textcolor{care}{\textbf{\faHandsHelping\ Care}}}
  & \textsc{Def:} Aversion to others' suffering; underlies kindness and nurturance. \\
  & \textsc{Ex:} ``My \textcolor{care}{heart breaks} seeing \textcolor{care}{children separated} from families at the border.'' \\
\midrule
\multirow{2}{*}{\textcolor{fairness}{\textbf{\faBalanceScale\ Fairness}}}
  & \textsc{Def:} Demand for equal treatment and reciprocity; underlies justice. \\
  & \textsc{Ex:} ``Everyone deserves \textcolor{fairness}{equal access} to healthcare \textcolor{fairness}{regardless} of income.'' \\
\midrule
\multirow{2}{*}{\textcolor{authority}{\textbf{\faCrown\ Authority}}}
  & \textsc{Def:} Deference to rank and tradition; underlies respect and duty. \\
  & \textsc{Ex:} ``\textcolor{authority}{Respect} your elders and follow \textcolor{authority}{traditional} values that built this nation.'' \\
\midrule
\multirow{2}{*}{\textcolor{loyalty}{\textbf{\faShield*\ Loyalty}}}
  & \textsc{Def:} Allegiance to the in-group over outsiders; underlies patriotism. \\
  & \textsc{Ex:} ``\textcolor{loyalty}{Stand with our troops} --- they sacrifice everything for our freedom.'' \\
\midrule
\multirow{2}{*}{\textcolor{sanctity}{\textbf{\faDove\ Sanctity}}}
  & \textsc{Def:} Reverence for the pure and untainted; underlies chastity and restraint. \\
  & \textsc{Ex:} ``\textcolor{sanctity}{Marriage} is \textcolor{sanctity}{sacred} and should be protected from secular corruption.'' \\
\bottomrule
\end{tabular}
\caption{The five moral foundations: brief definitions and illustrative social-media examples. Key moral cues highlighted in foundation color~\cite{skorskiMoralGapLarge2025}.}
\label{tab:mft-combined}
\end{table}

Moral Foundations Theory \citep{haidtWhenMoralityOpposes2007,graham2013moral} posits five innate moral intuitions --- Care, Fairness, Loyalty, Authority, Sanctity --- universal in kind, variable in weight; that variance is what drives ideological disagreement. A fixed, theory-derived label set maps naturally onto multi-label classification, arguably making MFT the moral-psychology framework best suited to computational treatment: a growing body of work fine-tunes language models to detect which foundations a text invokes \citep{hooverMoralFoundationsTwitter2020,tragerMoralFoundationsReddit2022,hoppExtendedMoralFoundations2021,preniqiMoralBERTFineTunedLanguage2024,zangari-etal-2025-me2,bullaLargeLanguageModels2025}. This research line trains on hard labels: annotator votes are collapsed once, by majority or any-annotator rule, upstream of the classifier. We depart in kind, not merely in degree --- not a hard label, not an ad hoc soft label (raw agreement fraction), but a soft label estimated as the posterior of a fully Bayesian consensus model, with calibrated rather than heuristic uncertainty.

Bayesian aggregation of noisy annotations has a long history, most notably Dawid--Skene \citep{dawidMaximumLikelihoodEstimation1979} and its extensions \citep{paunAggregatingLearningMultiple2021}, which our model builds on directly. Aleatoric/epistemic uncertainty decomposition is standard in Bayesian deep learning, but for model uncertainty over inputs, not annotator uncertainty over labels; we adapt it to aggregation. To our knowledge this has not been used to audit consensus-labeling heuristics in moral judgment. The paper is methodologically complementary to \citet{skorskiBayesianEvaluationLLMs2025}, which uses a similar Bayesian consensus model to compare LLM and human accuracy; here we extend the model to a fully Bayesian treatment and apply it to auditing consensus rules, entropy-based uncertainty analysis, and soft-label fine-tuning, rather than only to a pre-trained downstream predictor.

\section{Methods}\label{sec:method}

\subsection{Bayesian Consensus Model}\label{sec:bayesian-consensus}

\begin{figure}[h!]
\centering
\resizebox{\columnwidth}{!}{%
\begin{tikzpicture}
  \node[det]                            (alpha) {$\boldsymbol{\alpha}$};
  \node[latent, right=1.8cm of alpha]    (pi)    {$\boldsymbol{\pi}$};
  \node[det,    right=3.2cm of pi]       (beta)  {$\boldsymbol{\beta}_k$};
  \node[latent, right=1.8cm of beta]     (theta) {$\boldsymbol{\theta}_{jk}$};
  \node[latent, below=2.0cm of pi]       (z) {$z_i$};
  \node[obs,    below=2.0cm of theta]    (y) {$y_{ij}$};

  \edge {alpha} {pi};
  \edge {beta}  {theta};
  \edge {pi}    {z};
  \edge {theta} {y};
  \edge {z}     {y};

  \plate[inner sep=0.4cm] {annotator-block} {(beta)(theta)} {$j=1,\dots,J,\ k=1,\dots,K$};
  \plate[inner sep=0.3cm] {item-plate}      {(z)(y)}        {$i=1,\dots,N,\ j=1,\dots,J$};
\end{tikzpicture}%
}
\caption{Plate diagram of the Dawid--Skene model.}
\label{fig:dawid-skene-plates}
\end{figure}
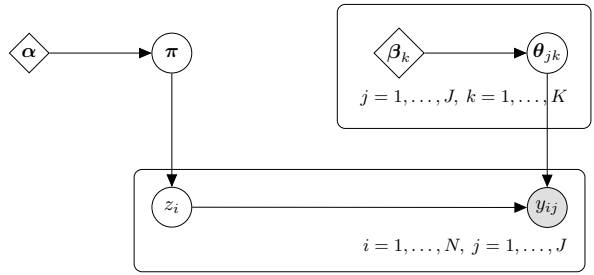

We adapt a Dawid--Skene-style model \citep{dawidMaximumLikelihoodEstimation1979} with weak Dirichlet priors, following the specification in \citet{skorskiBayesianEvaluationLLMs2025}. There are $N$ items, each with a true label drawn from $K$ possible classes with prevalence $\boldsymbol{\pi} \sim \mathrm{Dir}(\boldsymbol{\alpha})$. Each of $J$ annotators has a confusion matrix $\boldsymbol{\Theta}_j$, whose rows $\boldsymbol{\theta}_{jk} \sim \mathrm{Dir}(\boldsymbol{\beta}_k)$ encode a weak prior belief that annotators are more often right than wrong. For item $i$, given its annotations $\mathbf{y}_i$ (from up to $J$ annotators; annotations may be missing when not completed by an annotator or excluded in post-review, e.g. based on self-reports), the model yields a posterior 
\[
\Pr\{z_i = k \mid \mathbf{y}_i\} \propto \pi_k \prod_{j \in \mathcal{J}_i} \theta_{jk, y_{ij}},
\]
where $\mathcal{J}_i$ indexes the annotators who labelled item $i$.
The corresponding generative model is shown in plate notation in \Cref{fig:dawid-skene-plates}.

We first fit the model by MAP estimation via gradient ascent on the log-posterior (Adam, log-domain computation for numerical stability); then, to move from a point estimate to full posterior uncertainty, we approximate the posterior with a Laplace approximation --- a Gaussian centered at the MAP estimate with covariance given by the inverse negative Hessian of the log-posterior. We take this route because exact sampling via NUTS proved fairly slow; reassuringly, our epistemic--aleatoric entropy decomposition (\Cref{sec:entropy-decomposition}) shows epistemic uncertainty to be small relative to aleatoric, confirming post hoc that the MAP estimate was already a good approximation to the full posterior.
We refer to the resulting per-item probability as $p_i = \Pr\{z_i = 1 \mid \mathbf{y}_i\}$, the \emph{Bayesian consensus}; a GPU-optimized implementation in PyTorch is shared in \Cref{lst:dawid-skene}.

\begin{lstlisting}[style=pytorch, caption={PyTorch implementation of Moral Consensus model. \texttt{forward} returns per-item log-scores via a batched \texttt{embedding\_bag} recombine over ragged annotator sets; \texttt{log\_p} adds Dirichlet priors for MAP training.},
captionpos=b,
label={lst:dawid-skene},
]
import numpy as np
import torch
import torch.nn as nn
import torch.nn.functional as F
from torch.distributions import Dirichlet
from tqdm import tqdm

def annotations_to_bag(annotations, K):
    """(N,J) array with NaN for missing -> flat_ids, offsets for embedding_bag."""
    N, J = annotations.shape
    valid_mask = ~np.isnan(annotations)
    n_coords, j_coords = np.where(valid_mask)
    order = np.argsort(n_coords, kind='stable')       # group by item
    n_coords, j_coords = n_coords[order], j_coords[order]
    labels = annotations[valid_mask].astype(int)[order]

    flat_ids = torch.as_tensor(j_coords * K + labels, dtype=torch.long)
    counts = np.bincount(n_coords, minlength=N)
    offsets = torch.as_tensor(np.concatenate([[0], np.cumsum(counts)[:-1]]), dtype=torch.long)
    return flat_ids, offsets

class DawidSkene(nn.Module):
    def __init__(self, J, K, alpha):
        super().__init__()
        self.J, self.K = J, K
        self.pi_logits = nn.Parameter(torch.randn(K) * 0.1)
        init_theta = (0.6 * torch.eye(K) + 0.4 / K).expand(J, K, K).clone()
        self.theta_logits = nn.Parameter(torch.log(init_theta + 1e-8))

        self.register_buffer('alpha', torch.as_tensor(alpha, dtype=torch.float32))
        conf_alpha = torch.full((J, K, K), 0.4)
        conf_alpha.diagonal(dim1=-2, dim2=-1).fill_(0.6)
        self.register_buffer('conf_alpha', 3 * conf_alpha)

    @classmethod
    def from_annotations(cls, annotations, alpha, device='cpu'):
        K = int(np.nanmax(annotations)) + 1
        N, J = annotations.shape
        model = cls(J, K, alpha).to(device)
        flat_ids, offsets = annotations_to_bag(annotations, K)
        return model, flat_ids.to(device), offsets.to(device)

    def forward(self, flat_ids, offsets):
        """Returns unnormalized per-item log-scores over true class, shape (N, K)."""
        log_pi = F.log_softmax(self.pi_logits, dim=-1)
        log_theta = F.log_softmax(self.theta_logits, dim=-1)

        lt = log_theta.transpose(1, 2).reshape(self.J * self.K, self.K)
        lp = F.embedding_bag(flat_ids, lt, offsets, mode='sum')
        logits = lp + log_pi[None, :]
        return logits  # (N, K)

    def log_p(self, logits):
        """Marginal log-likelihood + log-prior, given logits from forward()."""
        log_pi = F.log_softmax(self.pi_logits, dim=-1)
        log_theta = F.log_softmax(self.theta_logits, dim=-1)
        pi, theta = log_pi.exp(), log_theta.exp()

        ll = torch.logsumexp(logits, dim=1).sum()
        ll = ll + Dirichlet(self.alpha).log_prob(pi)
        ll = ll + Dirichlet(self.conf_alpha).log_prob(theta).sum()
        return ll

    def posterior(self, logits):
        return F.softmax(logits, dim=-1)

    def fit(self, flat_ids, offsets, max_iter=200, lr=1e-2):
        opt = torch.optim.Adam(self.parameters(), lr=lr)
        pbar = tqdm(range(max_iter))
        for _ in pbar:
            opt.zero_grad()
            logits = self(flat_ids, offsets)
            loss = -self.log_p(logits)
            loss.backward()
            opt.step()
            pbar.set_postfix(loss=loss.item())
        return self
\end{lstlisting}

\subsection{Moral Entropy: Total, Aleatoric, and Epistemic}
For a binary foundation label, the total predictive uncertainty of item $i$ is the entropy of the consensus probability (the posterior mean over draws),
\[
H_{\text{tot}}(i) = -p_i \log p_i - (1-p_i)\log(1-p_i),
\]
which we call the item's \emph{Moral Entropy}. Given posterior draws $p_i^{(1)}, \dots, p_i^{(S)}$ --- here obtained by sampling from the Laplace approximation to the posterior (\Cref{sec:bayesian-consensus}) --- $H_{\text{tot}}$ decomposes as
\[
H_{\text{tot}}(i) = \underbrace{\tfrac{1}{S}\textstyle\sum_s H(p_i^{(s)})}_{\text{aleatoric}} + \underbrace{H_{\text{tot}}(i) - \tfrac{1}{S}\textstyle\sum_s H(p_i^{(s)})}_{\text{epistemic}},
\]
following the mutual-information decomposition standard in Bayesian deep learning. The aleatoric term is the uncertainty a fully-informed model would still report --- true, irreducible moral ambiguity. The epistemic term, non-negative by Jensen's inequality, reflects disagreement \emph{among} plausible models, e.g.\ from too few or too noisy annotators on an item --- uncertainty that more annotation could reduce. \Cref{sec:entropy-decomposition} reports this split empirically and finds epistemic uncertainty consistently negligible, so $H_{\text{tot}}$ is effectively aleatoric throughout.

\subsection{Grading Common Consensus Rules}
The Moral Entropy $H_{\text{tot}}(i)$ is maximized exactly at $p_i = 1/2$ --- the point of maximal ambiguity --- and thresholding there is also the Bayes-optimal decision under 0/1 loss, minimizing balanced error. This lets the Bayesian consensus double as ground truth: from $p_i$ we obtain a gold label $z_i^\star = \mathbf{1}[p_i \geq 1/2]$, against which we grade the discrete consensus/decision rules popular in prior work --- the any-annotator rule $g_i^{\text{any}} = \mathbf{1}[\sum_j y_{ij} \geq 1]$, majority vote $g_i^{\text{maj}} = \mathbf{1}[\bar y_i \geq 1/2]$, and the absolute two-vote rule $g_i^{\text{2v}} = \mathbf{1}[\sum_j y_{ij} \geq 2]$. Writing $g_i$ for the consensus rule under study, we grade $g$ against $p_i$ and $z_i^\star$ as follows.

$g$ can also be compared to the continuous consensus $p_i$ via cross-entropy/KL, the Brier score's reliability--resolution--uncertainty decomposition, and Expected Calibration Error; these arise as Bregman divergences under different convex generators and aggregation norms, and we report them in full in the supplementary material\footnote{\url{https://github.com/maciejskorski/moral-entropy/}}.

In the main text we report the zeroth-order member of this family: hard disagreement against the Bayes-optimal gold label,
\[
\mathrm{FPR} = \Pr[g_i{=}1 \mid z_i^\star{=}0], \quad \mathrm{FNR} = \Pr[g_i{=}0 \mid z_i^\star{=}1].
\]
This is informative and assumption-free, though blind to the \emph{severity} of a disagreement, which the continuous diagnostics in the supplement restore.
\section{Data}

We evaluate on three MFT-annotated corpora, summarized in \Cref{tab:corpora}. \textbf{MFTC} \citep{hooverMoralFoundationsTwitter2020} provides roughly $125{,}000$ annotations from 23 trained annotators across $33{,}858$ tweets spanning seven discourse domains (\#BlackLivesMatter, 2016 U.S. Election, \#MeToo, Hurricane Sandy, Baltimore protests, \#AllLivesMatter, and Davidson's hate-speech corpus). \textbf{MFRC} \citep{tragerMoralFoundationsReddit2022} provides 17,886 Reddit posts across three subcorpora --- Everyday Morality (r/AmItheAsshole), U.S. politics, and French politics --- capturing more colloquial, informally structured moral reasoning than MFTC's tweets. \textbf{eMFD} \citep{hoppExtendedMoralFoundations2021} is by far the largest of the three (54,883 items here) but topically narrower, drawn entirely from GDELT-indexed news-article paragraphs; we ourselves group these into subcorpora by GDELT theme tags (Physical Violence, Asymmetric Threats, Internal Unrest, Civil Liberties, plus an ``Untagged'' residual category) for this study, to enable cross-domain comparison to the social-media corpora.
 
\begin{table}[h!]
\centering
\tiny
\setlength{\tabcolsep}{4pt}
\begin{tabular}{@{}llrrrrrrr@{}}
\toprule
Corpus & Subcorpus & Au\% & Ca\% & Fa\% & Lo\% & Sa\% & \#Ann. & $N$ \\
\midrule
\multirow{3}{*}{MFRC}
  & Everyday Morality & 10.4 & 37.4 & 25.5 & 11.7 & 13.5 & 3.00 &  5{,}366 \\
  & US Politics       & 19.7 & 29.6 & 38.3 &  7.7 &  8.4 & 2.99 &  5{,}351 \\
  & French Politics   & 25.4 & 16.0 & 26.0 & 13.1 &  8.0 & 3.00 &  7{,}169 \\
\cmidrule{2-9}
  & \multicolumn{1}{l}{\textit{MFRC total}} & & & & & & & 17{,}886 \\
\midrule
\multirow{7}{*}{MFTC}
  & ALM       & 20.9 & 61.6 & 49.8 & 26.3 & 16.9 & 3.03 &  4{,}326 \\
  & BLM       & 21.3 & 56.6 & 49.5 & 27.2 & 20.8 & 4.92 &  5{,}117 \\
  & Baltimore & 31.8 & 27.1 & 31.4 & 42.3 &  9.0 & 3.32 &  5{,}190 \\
  & Davidson  & 33.0 & 11.5 & 11.2 & 11.6 & 12.3 & 3.78 &  4{,}873 \\
  & Election  & 17.5 & 38.0 & 34.4 & 22.8 & 30.4 & 3.88 &  5{,}050 \\
  & MeToo     & 65.7 & 33.3 & 43.9 & 41.0 & 52.8 & 3.65 &  4{,}711 \\
  & Sandy     & 45.6 & 60.9 & 30.3 & 42.7 & 15.0 & 3.03 &  4{,}591 \\
\cmidrule{2-9}
  & \multicolumn{1}{l}{\textit{MFTC total}} & & & & & & & 33{,}858 \\
\midrule
\multirow{5}{*}{eMFD}
  & Physical Violence  & 24.3 & 24.5 & 21.1 & 22.4 & 18.3 & 1.20 & 30{,}261 \\
  & Asymmetric Threats & 26.1 & 22.2 & 24.1 & 26.8 & 18.0 & 1.44 &  6{,}622 \\
  & Internal Unrest    & 25.0 & 26.7 & 29.4 & 26.6 & 23.1 & 1.83 &  5{,}758 \\
  & Civil Liberties    & 26.0 & 23.6 & 26.1 & 22.2 & 18.3 & 1.40 & 10{,}334 \\
  & Untagged           & 21.6 & 30.3 & 19.4 & 17.8 & 18.7 & 1.11 &  1{,}908 \\
\cmidrule{2-9}
  & \multicolumn{1}{l}{\textit{eMFD total}} & & & & & & & 54{,}883 \\
\bottomrule
\end{tabular}
\caption{Corpora, subcorpora, and per-foundation prevalence (\% of total texts). Au=authority, Ca=care, Fa=fairness, Lo=loyalty, Sa=sanctity. Ann.=mean number of annotations per text.}
\label{tab:corpora}
\end{table}

\section{Results}

We fit the Bayesian consensus model per foundation and per corpus (MFTC, MFRC, eMFD), compute $p_i$ (Bayesian consensus) alongside heuristic consensus schemes $g_i$ (any-annotator, two-vote, and majority) for every item, then study uncertainty and biases with entropy per discourse domain.

\subsection{Audit of Moral Consensus Rules}





\begin{figure}[h!]
\centering
\includegraphics[width=0.98\columnwidth]{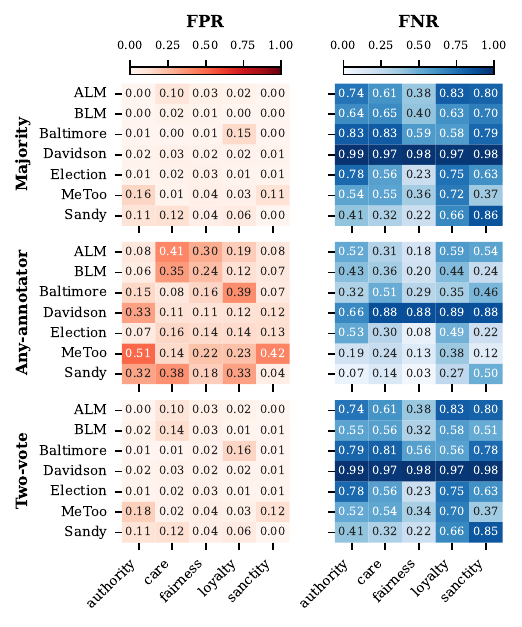}
\caption{MFTC: per-foundation FPR/FNR of the majority, any-annotator, and two-vote rules against the Bayesian consensus, across the seven discourse domains and five MFT foundations.}
\label{fig:fpr-fnr-mftc}
\end{figure}

Pooling across corpora, the any-annotator rule shows $\mathrm{FPR}=30.6\%$ against the Bayesian consensus and $\mathrm{FNR}=0.0\%$: since $g_i{=}0$ requires \emph{every} annotator to vote negative, and the confusion-matrix prior is diagonal-dominant, an all-negative vote pattern almost never yields a posterior above 0.5 --- the asymmetry is close to structurally guaranteed by the OR-rule, not an empirical coincidence. The $30.6\%$ false-positive rate, in contrast, is genuine and substantial: \textbf{roughly a third of items flagged positive by the naive rule are \emph{not} supported by the calibrated Bayesian consensus}. It also varies sharply across discourse domains --- per-foundation $\mathrm{FPR}$ ranges from $12.8\%$ (Election) to $30.4\%$ (MeToo) and $\mathrm{FNR}$ from $20.2\%$ (Sandy) to $83.8\%$ (Davidson); MeToo and Sandy are in fact the only two domains where $\mathrm{FPR}$ still exceeds $\mathrm{FNR}$, preserving the pooled false-positive-dominant pattern, while the other five domains (ALM, BLM, Baltimore, Davidson, Election) invert it, with $\mathrm{FNR}$ the larger error --- so \textbf{item-level moral ambiguity and aggregation-rule bias are independent axes} (\Cref{fig:fpr-fnr-mftc}).
 
\begin{figure}[h!]
\centering
\includegraphics[width=0.98\columnwidth]{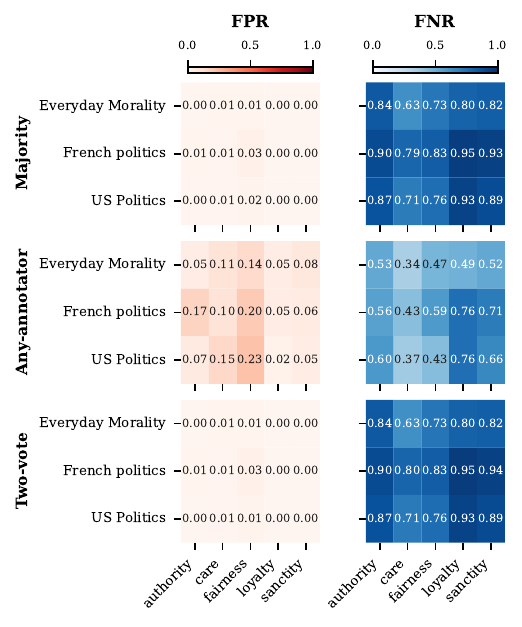}
\caption{MFRC: per-foundation FPR/FNR of the three rules against the Bayesian consensus, across Everyday Morality, French Politics, and US Politics.}
\label{fig:fpr-fnr-mfrc}
\end{figure}

No common vote-count rule is well calibrated. Breaking $\mathrm{FPR}/\mathrm{FNR}$ out by the five MFT foundations, and adding two stricter rules --- an absolute two-vote threshold and a per-item majority vote --- shows any-annotator's near-zero pooled $\mathrm{FNR}$ does not survive at the foundation level (mean $\mathrm{FPR}/\mathrm{FNR}$: $19.9\%/38.9\%$ on MFTC, $10.2\%/54.8\%$ on the three MFRC domains), while majority and two-vote invert the failure --- nearly eliminating false positives ($3.5\%$ and $4.2\%$ on MFTC, under $1\%$ on MFRC) but missing most true positives instead (mean $\mathrm{FNR}$ $63$--$65\%$ on MFTC, $\approx83\%$ on MFRC). The two strict rules are near-identical only where annotation is thin --- on MFRC $13$ of $15$ $\mathrm{FNR}$ cells coincide, since at $3.00$ annotations per item a majority \emph{is} two votes --- and separate where items carry more annotators (BLM, $4.92$ per item: $\mathrm{FNR}$ $0.64$ vs.\ $0.55$) (\Cref{fig:fpr-fnr-mfrc}).
 
\begin{figure}[h!]
\centering
\includegraphics[width=0.98\columnwidth]{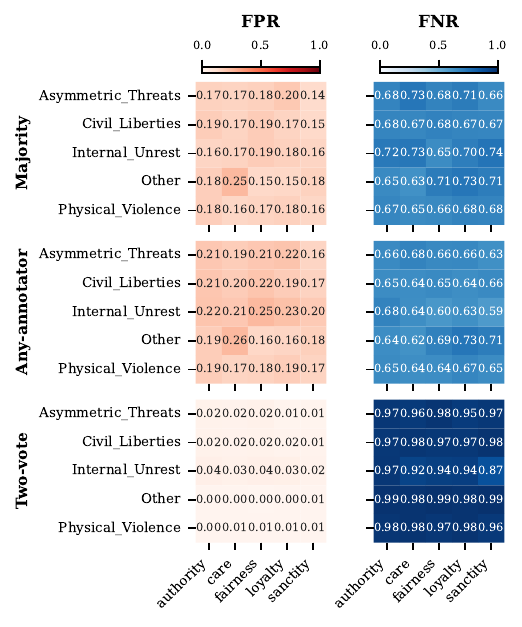}
\caption{eMFD: per-foundation FPR/FNR of the three rules against independent gold labels, across the five topic categories.}
\label{fig:fpr-fnr-emfd}
\end{figure}

An independent check against eMFD's expert gold labels, rather than our own posterior, confirms the pattern and exposes one rule's failure outright: two-vote becomes nearly degenerate (mean $\mathrm{FPR}=1.5\%$, $\mathrm{FNR}=96.6\%$), since most eMFD items never reach two votes at all. \textbf{The same absolute threshold that looked comparatively clean on MFTC is nearly useless on eMFD} (\Cref{fig:fpr-fnr-emfd}) --- no fixed vote-count rule generalizes across annotation designs, and which rule looks ``safe'' is itself dataset-dependent.
 
\subsection{Soft-Label Fine-Tuning}

Instead of a collapsed hard label, we fine-tune classifiers on MFTC directly on the calibrated posterior $p_i$ as a soft target. Soft-label models beat identical architectures trained on aggregated voting labels by 2--3\% accuracy, consistently across foundations. The gain tracks the argument above: $p_i$ never forces false confidence onto an ambiguous item, so the model isn't trained to be certain where the annotators weren't. Training is stable and fast --- \Cref{tab:soft-label-auc} reports final ROC-AUC against the hard label, plateauing within the first epoch; Care converges highest at 0.955.

\begin{table}[h!]
\centering
\tiny
\begin{tabular}{lr}
\toprule
Foundation & ROC-AUC vs.\ Hard \\
\midrule
Sanctity   & 0.950 \\
Authority  & 0.944 \\
Loyalty    & 0.934 \\
Care       & 0.955 \\
Fairness   & 0.949 \\
\bottomrule
\end{tabular}
\caption{Soft-label fine-tuning on MFTC: final ROC-AUC of the soft-label model against the hard (aggregated-vote) label, per foundation. Reported as an agreement check; the 2--3\% accuracy comparison is given in the text.}
\label{tab:soft-label-auc}
\end{table}

\subsection{Entropy Decomposition}\label{sec:entropy-decomposition}

\Cref{tab:entropy-decomp} shows that total posterior entropy is \textbf{almost entirely aleatoric}. In every corpus/foundation cell the epistemic term is one to three orders of magnitude smaller than the aleatoric term: it contributes \textbf{under $1\%$} of total entropy in MFTC (mean $0.3\%$) and MFRC (mean $0.3\%$), and \textbf{never exceeds $6\%$} even in eMFD, the corpus with the fewest annotators per item. The model's parameters are therefore tightly constrained by the data --- the posterior is not merely reflecting our own uncertainty about the model itself. So the \textbf{entropy reported is real moral disagreement}, not an artifact of model underfitting or poor data quality. This also justifies the posterior approximation used here: with epistemic uncertainty this small, the posterior is well-approximated by a point estimate, so the cost of exact MCMC sampling would buy little while a Laplace approximation gives the same reassurance at a fraction of the cost --- and scales.

\begin{table}[h!]
\centering
\small
\setlength{\tabcolsep}{4pt}
\resizebox{\linewidth}{!}{
\begin{tabular}{lrrrrrrrrrr}
\toprule
 & \multicolumn{2}{c}{Authority} & \multicolumn{2}{c}{Care} & \multicolumn{2}{c}{Fairness} & \multicolumn{2}{c}{Loyalty} & \multicolumn{2}{c}{Sanctity} \\
\cmidrule(lr){2-3}\cmidrule(lr){4-5}\cmidrule(lr){6-7}\cmidrule(lr){8-9}\cmidrule(lr){10-11}
Corpus & Ale & Epi & Ale & Epi & Ale & Epi & Ale & Epi & Ale & Epi \\
\midrule
\multicolumn{11}{l}{\textit{MFRC}} \\
\midrule
Everyday Morality & 0.0670 & 0.0002 & 0.0542 & 0.0002 & 0.0760 & 0.0002 & 0.0661 & 0.0002 & 0.0587 & 0.0002 \\
French politics & 0.1385 & 0.0003 & 0.0348 & 0.0001 & 0.0856 & 0.0002 & 0.1420 & 0.0003 & 0.0520 & 0.0002 \\
US Politics & 0.1126 & 0.0002 & 0.0531 & 0.0002 & 0.0902 & 0.0002 & 0.0928 & 0.0002 & 0.0588 & 0.0002 \\
\midrule
\multicolumn{11}{l}{\textit{MFTC}} \\
\midrule
ALM & 0.0771 & 0.0001 & 0.0799 & 0.0001 & 0.1015 & 0.0001 & 0.0790 & 0.0001 & 0.0485 & 0.0001 \\
BLM & 0.1066 & 0.0002 & 0.0824 & 0.0002 & 0.0944 & 0.0002 & 0.0917 & 0.0002 & 0.0307 & 0.0001 \\
Baltimore & 0.0969 & 0.0004 & 0.0586 & 0.0002 & 0.0843 & 0.0003 & 0.0643 & 0.0002 & 0.0226 & 0.0002 \\
Davidson & 0.0349 & 0.0002 & 0.0729 & 0.0002 & 0.0203 & 0.0001 & 0.0713 & 0.0002 & 0.1321 & 0.0002 \\
Election & 0.0773 & 0.0001 & 0.0551 & 0.0001 & 0.0384 & 0.0001 & 0.0663 & 0.0001 & 0.0414 & 0.0002 \\
MeToo & 0.1257 & 0.0006 & 0.0421 & 0.0003 & 0.0617 & 0.0004 & 0.0953 & 0.0003 & 0.0473 & 0.0003 \\
Sandy & 0.0697 & 0.0002 & 0.0667 & 0.0002 & 0.0333 & 0.0002 & 0.0734 & 0.0002 & 0.0552 & 0.0002 \\
\midrule
\multicolumn{11}{l}{\textit{eMFD}} \\
\midrule
Asymmetric Threats & 0.1168 & 0.0031 & 0.0517 & 0.0015 & 0.0609 & 0.0039 & 0.1097 & 0.0026 & 0.0387 & 0.0018 \\
Civil Liberties & 0.1088 & 0.0035 & 0.0428 & 0.0011 & 0.0670 & 0.0025 & 0.0860 & 0.0025 & 0.0334 & 0.0013 \\
Internal Unrest & 0.1063 & 0.0036 & 0.0503 & 0.0015 & 0.0680 & 0.0034 & 0.1012 & 0.0046 & 0.0620 & 0.0030 \\
Physical Violence & 0.1055 & 0.0026 & 0.0446 & 0.0010 & 0.0511 & 0.0018 & 0.0921 & 0.0020 & 0.0352 & 0.0015 \\
Untagged & 0.0821 & 0.0020 & 0.0504 & 0.0012 & 0.0527 & 0.0012 & 0.0798 & 0.0017 & 0.0327 & 0.0011 \\
\bottomrule
\end{tabular}
}
\caption{Aleatoric and epistemic components of posterior entropy (nats) by corpus and moral foundation.}
\label{tab:entropy-decomp}
\end{table}

\subsection{Outliers in Moral Judgment}

Moral Entropy can also localize specific items and annotators worth closer inspection. Among the 100 highest-entropy Care items in BLM, disagreement concentrates on \textbf{hostile political rhetoric} (\textit{``terrorists,''} \textit{``TERRORISM,''} \textit{``shoot tyrants''}) rather than on compassion-laden content. Disagreement rates against the BLM majority range from 7.6\% (A04) to 31.3\% (A02) --- some annotators are simply more discordant than others.

\Cref{tab:annotator-switching} shows five representative items. \textbf{Almost no annotator plays a fixed role}: A00 agrees with the majority on all five and A03 flags Care on all five, while A04 and A01 each dissent twice --- but on different items, so neither is a consistent outlier. Most tellingly, A02, the most discordant annotator corpus-wide (31.3\%), agrees with the majority on four of these five items and dissents alone only on the least ambiguous one --- a wish for someone's death.

This tracks a known difficulty in moral psychology: violent, dehumanizing rhetoric resists clean foundation assignment. \citet{kennedyMoralLanguageHate2023} find that hateful language mostly invokes Purity and Loyalty/Authority, implicating Care only when an explicit harmful act is described. The items in \Cref{tab:annotator-switching} sit exactly on that boundary --- harm-adjacent language, no concrete act --- which is precisely where annotators diverge. \textbf{Moral Entropy recognizes the uncertainty of this class accurately, instead of silently burying it in a majority vote.}


\begin{table}[h!]
\centering
\tiny
\setlength{\tabcolsep}{1.5pt}
\begin{annotatorbox}
\begin{tabular}{p{0.52\columnwidth}cccccc}
\toprule
\textbf{Text} & \textbf{A00} & \textbf{A01} & \textbf{A02} & \textbf{A03} & \textbf{A04} & \textbf{Maj.} \\
\midrule
``Sick of obama \& his leftist sycophant propagandists @washingtonpost supporting \#blacklivesmatter terrorists.'' &
0 & 0 & 0 & \textbf{1} & \textbf{1} & 0 \\
\addlinespace
``\#I94ShutDown ignorant idiots promoting ANARCHY over the death of CRIMINALS.. \#BLM is TERRORISM'' &
0 & 0 & 0 & \textbf{1} & \textbf{1} & 0 \\
\addlinespace
``\#AllLivesMatter sick of the racism bullshit. Stop keeping it alive by putting a certain race in the spotlight.'' &
\textbf{1} & 0 & \textbf{1} & \textbf{1} & \textbf{1} & 1 \\
\addlinespace
``\#RightWingLogic u need guns to shoot tyrants, but always obey the police!'' &
0 & \textbf{1} & 0 & \textbf{1} & 0 & 0 \\
\addlinespace
``i hope this nigger suffers as he dies he deserves to suffer and die in pain'' &
\textbf{1} & \textbf{1} & 0 & \textbf{1} & \textbf{1} & 1 \\
\bottomrule
\end{tabular}
\end{annotatorbox}
\caption{Annotator votes on Care for five representative high-entropy BLM items. No annotator occupies a fixed ``permissive'' or ``conservative'' role: the same annotators switch sides depending on item wording. \textbf{Content warning:} the final item contains a racial slur, quoted verbatim from the corpus.}
\label{tab:annotator-switching}
\end{table}

\section{Discussion}
These results motivate three changes to moral inference pipelines.
\begin{itemize}
\item \textbf{Report entropy.} Entropy should accompany any consensus label, decomposed into aleatoric and epistemic components (\Cref{sec:entropy-decomposition}), so a ``hard'' corpus can be diagnosed as genuinely contested versus under-annotated --- and so any heuristic consensus rule already in use can be audited against it, since a rule can look reasonable in isolation while hiding a large, systematic, corpus-dependent bias.
\item \textbf{Train on entropic (soft) labels.} Training directly on the posterior probability --- an entropic, uncertainty-aware label --- in place of a collapsed hard one, yields consistent 2--3\% accuracy gains (\Cref{tab:soft-label-auc}), driven by the greater consistency of a calibrated soft target relative to a heuristic that lacks uncertainty.
\item \textbf{Posterior entropy is cheap to approximate.} Epistemic uncertainty is a small, near-negligible share of total entropy across corpora (\Cref{tab:entropy-decomp}), meaning the confusion-matrix parameters are tightly identified by the data. A Laplace approximation around the MAP fit suffices in place of costly full MCMC --- making Bayesian consensus scalable.
\end{itemize}
 
Together, these three points reframe annotator disagreement as a resource rather than a nuisance to vote away: cheap to model, informative to report, and directly exploitable at training time, with no accuracy cost --- soft labels beat the hard labels they were built from. None of this machinery is specific to Moral Foundations Theory; entropy audit can be applied to any values- or judgment-annotation pipeline before a hard-label dataset ships.

\section*{Limitations}
\label{sec:limitations}

\paragraph{Aleatoric/epistemic decomposition.} The decomposition reported in \Cref{tab:entropy-decomp} rests on a Laplace approximation around the MAP fit, not on full posterior draws. We attempted a direct validation via NUTS sampling, but it suffered from slow mixing (effective sample size as low as 5 out of 300 draws even after tuning burn-in and trajectory depth, likely due to strong correlations in the high-dimensional confusion-matrix parameter space); on the small number of items where MCMC did yield usable draws, the resulting decomposition matched the Laplace approximation, which is why we report the latter as our primary uncertainty decomposition. A systematic validation, via a reparameterized sampler or an ensemble of independently seeded MAP fits, is a natural extension we leave to future work.

\paragraph{Cultural and annotator-pool bias.} Our consensus model treats disagreement as informative rather than as noise to be voted away, but the annotator pool itself remains a source of unmodeled bias: annotators for MFTC (and MFRC, eMFD) were not sampled to represent diverse cultural or ideological perspectives, so even a well-calibrated consensus over this pool could still encode a comparatively narrow (plausibly WEIRD --- Western, Educated, Industrialized, Rich, Democratic) view of which cues count as morally salient. This risk is partly mitigated by the fact that annotators were trained against a shared codebook rather than left to apply unconstrained personal judgment: training standardizes what counts as a flag-worthy cue across annotators, which is exactly the regularity the consensus model relies on to be meaningful in the first place --- though it cannot correct for a codebook, or an annotator pool, that is itself culturally narrow. A second, complementary mitigation comes from recent translation research: \citet{skorski_moral_2026} shows that moral semantics are largely preserved under careful LLM-based translation (EN$\to$PL, AUROC gaps of 0.01--0.02 across foundations), pointing to a practical route for testing --- and eventually correcting --- this bias without new from-scratch annotation: translate an existing English corpus and re-run the Moral Entropy audit on it directly. We leave this cross-lingual test to future work.

\paragraph{Scope of the fine-tuning result.} The soft-label accuracy gains reported above are demonstrated on a subset of corpora/foundations and a single classifier architecture; generalization across architectures and the full label set is untested and left to future work.



\bibliography{citations}

\end{document}